\documentclass{article}

\usepackage{arxiv}

\usepackage[utf8]{inputenc} 
\usepackage[T1]{fontenc}    
\usepackage{hyperref}       
\usepackage{url}            
\usepackage{booktabs}       
\usepackage{amsfonts}       
\usepackage{nicefrac}       
\usepackage{microtype}      
\usepackage{lipsum}		
\usepackage{graphicx}
\usepackage{natbib}
\usepackage{doi}

\title{Using Adversarial Debiasing to remove bias from word embeddings}

\author{ {\hspace{1mm}Dana Kenna}\\
	\texttt{dkenna99@gmail.com} \\
}

\renewcommand{\shorttitle}{\textit{arXiv} Template}

\begin{document}
\maketitle

\begin{abstract}
Word Embeddings have been shown to contain the societal biases present in the original corpora. Existing methods to deal with this problem have been shown to only remove superficial biases. The method of \emph{Adversarial Debiasing} was presumed to be similarly superficial, but this is was not verified in previous works. Using the experiments that demonstrated the shallow removal in other methods, I show results that suggest \emph{Adversarial Debiasing} is more effective at removing bias and thus motivate further investigation on the utility of \emph{Adversarial Debiasing}. 

\end{abstract}

\section{Introduction}

Word Embeddings are used in a variety of Natural Language Processing (NLP) models to capture information about the relationships between words within a corpus. These embeddings are susceptible to inadvertently capturing social biases as inherent features of the words themselves. Consider a corpus about the first 46 Presidents of the United States of America. word2vec, a popular algorithm for efficient training of word embeddings developed by \cite{w2v}, learns representations of words from their neighbouring words in the text, and thus in the training process the word "President" would be closely linked to male-gendered words. This relationship is not capturing information about the presidency itself, which is not gendered in any way, but is instead capturing information about the demographics of past presidents. This presents an issue with the widespread usage of Word Embeddings built upon biased corpora. 

In this paper, I demonstrate results that show \emph{Adversarial Debiasing} can more effectively remove bias when compared to previous debiasing algorithms. These experiments, defined by \cite{gonen}, previously indicated that other debiasing techniques did not fully remove all forms of bias. The adversarially debiased embedding has notably improved performance on all of these experiments, indicating it has more effectively removed the more complex forms of bias. My results indicate the need for further exploration of the utility of \emph{Adversarial Debiasing}.

\section{Related Work}

In a previous work, \cite{bobulaski} provided an example of biased embeddings by showing that word embeddings that had been trained using word2vec on Google News Data \footnotemark resolved the analogy "man is to computer programmer as woman is to x", with "x = homemaker".In a literature review surveying 146 papers that analysed bias in NLP systems \cite{nlplit1} provided a number of recommendations to mitigate the potential for these systems to enact harm and perpetuate social biases. They argued for the importance of treating representational harms within a system as harmful in their own right, rather than only examining them as they affect downstream tasks.

\footnotetext{ The embeddings are available at \url{https://code.google.com/archive/p/word2vec/}. They were trained on part of the Google News Dataset (around 100 billion words) and contains 300 dimensional vectors for 3 million words and phrases}
In a previous work, \cite{bobulaski} presented a technique that removed the gender direction from the embedding. by taking gendered pairs such as ("he", "she") and transforming the vector space such that each word not in a gendered pair is equally close to both words in each gendered pair within the vector space \footnotemark. Taking their earlier example, this leads to "Computer Programmer" being as close to "he" as it is to "she". The same is also true for "Homemaker", and is true for each possibly gendered pair. The debiased embedding was shown to produce analogies that were less likely to be identified as stereotypical by Amazon Mechanical Turk crowd workers.

\footnotetext{This will be referred to throughout as the \emph{Vector Space Transformation}}
A recent work by \cite{gonen} argued this method did not effectively remove the bias, but instead mostly hid it. Whilst the space had been transformed, it relied on a particular definition of bias - as a words projection on the "gender direction". They claim bias is more profound and systemic, and reducing the projection of words on the direction is merely hiding the bias. Words like "math" and "delicate" are in principle gender neutral, but have strong gendered associations which will be present in the neighbouring words. Words biased in the same direction are still very likely to be clustered together after the transformation, even though their link to the explicitly gendered words has been removed.

\cite{gonen} used Principal Component Analysis (PCA), K-Means Clustering and Bias-By-Neighbours to demonstrate the superficial nature of this debiasing. Using a PCA, it can be observed that the most biased words along the gender direction are still clustered together after the transformation is applied. Using K-means clustering, they showed that when given the transformed embeddings of the most biased words they could still be accurately sorted into gendered clusters. 
\begin{figure}
    \centering
    \includegraphics[width=0.4\textwidth]{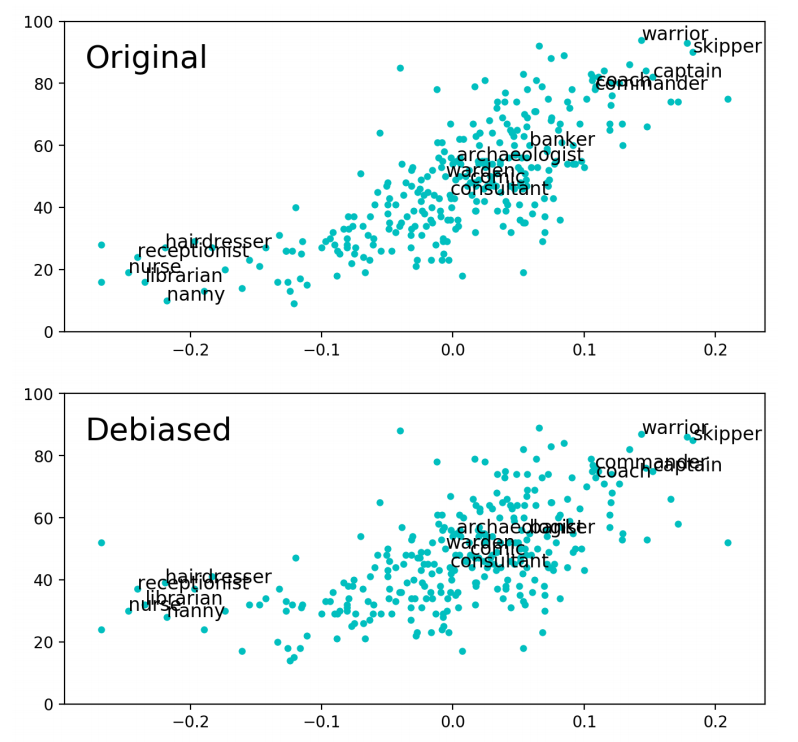}
    \caption{Figures courtesy of \cite{gonen} . It shows the the original bias projection on the X axis, plotted against the bias-by-neighbours - the percentage of neighbouring words that were previously biased male. We observe little change and bias projection can still be predicted from the neighbours, even after debiasing using the \emph{Vector Space Transformation}. Specific occupations are marked on the graph.}
    \label{fig:2bbn}
\end{figure}

They also showed that after transformations, the percentage of neighbouring words that had been previously biased male on the gender direction (the bias-by-neighbours) was strongly correlated with a words original bias projection on the gender direction. The correlation changes from 0.741 for the Non-Debiased embedding, to 0.686 for the \emph{Vector Space Transformation} embedding. The relationship is shown in the graphs in Figure \ref{fig:2bbn}. The original bias projection can be predicted from the relationships with words that have been preserved through the debiasing process. They concluded that whilst the technique had removed the bias on the gender projection, the debiasing was mostly superficial.

\emph{Adversarial Debiasing} is a debiasing technique introduced by \cite{advdebias} built upon the work of \cite{gans} on \emph{Generative Adversarial Networks}. \emph{Adversarial Debiasing} introduces an additional layer to an existing training process. Information from the original network is given to an adversary, which is tasked with predicting some protected characteristic. The loss gradients generated by this adversary are used to shape the training of the network, such that the network is unable to learn information that would aid the adversary in predicting the protected characteristic. This forces the network to attempt to find a new optima for the loss function which does not utilise the protected characteristic, thus learning a fairer embedding that does not rely upon this characteristic.

\cite{advdebias} demonstrated the utility of their technique for word embeddings, using the gender definition provided by \cite{bobulaski} to learn a transformation for an embedding that produced less biased analogies. Other works have used the technique, such as \cite{compasadv} which used \emph{Adversarial Debiasing} to remove bias from the COMPAS model for inmate recidivism . In this case, the protected variable was race. Their adversarial model significantly reduced the False Positive and False Negative gap between White and Black inmates.

In their evaluation of the \emph{Vector Space Transformation}, \cite{gonen} acknowledge \emph{Adversarial Debiasing} as another work in the spirit. They observe that the technique uses the same definition of bias as used in \cite{bobulaski} which they suggested would lead to similarly superficial removal of bias.

The \emph{Vector Space Transformation} was shown to preserve aspects of gender bias after debiasing but there is a distinction between the post-training \emph{Vector Space Transformation} of \cite{bobulaski}, and the technique of \emph{Adversarial Debiasing} which involves retraining the network. \cite{gonen} note in their conclusion that gender-direction is a way of measuring the gender-association, but does not determine it. \emph{Adversarial Debiasing} may result in a more complex transformation guided by the gender-direction, but not determined by it.

\section{Experimental Setup}

The assumption made by \cite{gonen} was that \emph{Adversarial Debiasing} and the \emph{Vector Space Transformation} would yield similar results for the bias experiments namely PCA, K-Means clustering and Bias-By-Neighbours correlation with the Original Bias Projection. The PCA experiment consists of visual examination of the PCA produced by the embeddings of the 500 most biased words for both male and female bias \cite{gonen} observed that clusters spaced out, but still were clearly distinct as shown in Figure \ref{PCAGON}. Even though the gender bias was removed, the most biased words are still very far apart in the space.

The K-Means clustering experiment takes the 500 most biased words in either direction, and attempts to cluster them into 2 clusters. The accuracy indicates how much the the biased distanced between words has been preserved through the debiasing process.

The Bias-By-Neighbours of a word is measured as the percentage of the 100 nearest neighbour words which were previously (prior to debiasing) biased in the male direction. This can be plotted against the original bias projection of the word to give a visual indication of the relationship between the two as shown in Figure \ref{fig:2bbn}, and a Pearson Correlation can be used to give a value for how well the original bias projection can be recovered from the neighbours after debiasing.

The original results were from the Google News Vectors. It contains embeddings for over 3 million words represented across 300 dimensions, and both \cite{bobulaski} and \cite{gonen} used the 50,000 most frequent words. These were filtered to only include words with less than 20 lower case characters, resulting in 26,377 remaining words.

This presents a huge training overhead and I used a significantly smaller corpus in my experiments. Whilst this does limit the usefulness of my results in terms of making conclusions about the overall utility of \emph{Adversarial Debiasing}, it does allow for differences between debiasing techniques to be observed. I used the text of 'Metamorphosis' \footnotemark as the corpus. When a minimum count of 2 is enforced (words that only appear once in the corpus are not embedded, as there is not enough information about them) this results in 1312 unique words embedded in a 300 dimensional vector space. I also used the word2vec algorithm to train this embedding, and also used 300 dimensions. Due to the smaller dataset, I took the 200 most biased words for the PCA and K-Means clustering experiments rather than the 500 used by \cite{gonen}.

\footnotetext{This was obtained from \url{https://www.gutenberg.org/}}

I used the word2vec algorithm to train a network for 750 epochs, using Metamorphosis as the corpus. This provided the base embedding, which we can attempt to debias. I implemented the linear transformation shown in \cite{bobulaski}, using just ("he", "she") as the gender pair from which the gender direction is learnt producing an embedding debiased with the \emph{Vector Space Transformation}.

I adapted my implementation from \cite{advdebias}. The word embedding experiment was adapted so that the embeddings were being updated, rather than a linear transformation being learned.In my implementation the adversary receives the embedding of the target word and attempt to predict the projection along the gender subspace. The gender subspace is defined in the same way as in the \emph{Vector Space Transformation}, using ("he","she") as the sole gender pair. The adversary would be able to improve its own weights freely as it is given pairs (X,Y). X is the 300 dimension word embedding, and Y is that words projection along the gender subspace. Every 10 epochs, the gender subspace projections are recalculated and updated. The network was trained for a further 750 epochs. The loss function was updated to avoid the network learning in a way that would aid the adversary in predicting the gendered projection. I used the parameters of alpha=1 and learning rate = 0.0001 \footnotemark.

\footnotetext{I repeated the experiment with other configurations of parameters, and observed that whilst some paramterisations were more effective than others all produced results which improved upon those resulting from the linear transformation}

Implementing the three experiments previously performed by \cite{gonen} on the smaller dataset yielded similar results from their work which suggests that a smaller dataset is able to exhibit similar trends to a larger one.

\section{Results}

\subsection{Comparison of the \emph{Vector Space Transformation} on a smaller dataset}

\begin{table}[htbp]
\resizebox{\textwidth}{!}{\begin{tabular}{lll}
Embedding                                                 & K-Means Accuracy & Bias-By-Neighbours Pearson Correlation \\
Non-Debiased (\cite{gonen})                               & 99.9\%           & 0.741                                  \\
Debiased with \emph{Vector Space Transformation} (\cite{gonen})  & 92.5\%           & 0.686                                  \\
Non-Debiased (My Embedding)                               & 99.4\%           & 0.884                                  \\
Debiased With  \emph{Vector Space Transformation} (My Embedding) & 88.9\%           & 0.6925                                 
\end{tabular}}
\caption{ A table showing the different results produced by applying the \emph{Vector Space Transformation}. Results are shown for the results found by \cite{gonen} and the results when testing on my own smaller dataset}
\label{tab:gonencomp}
\end{table}

\begin{figure}
    \centering
    \includegraphics[width=0.4\textwidth]{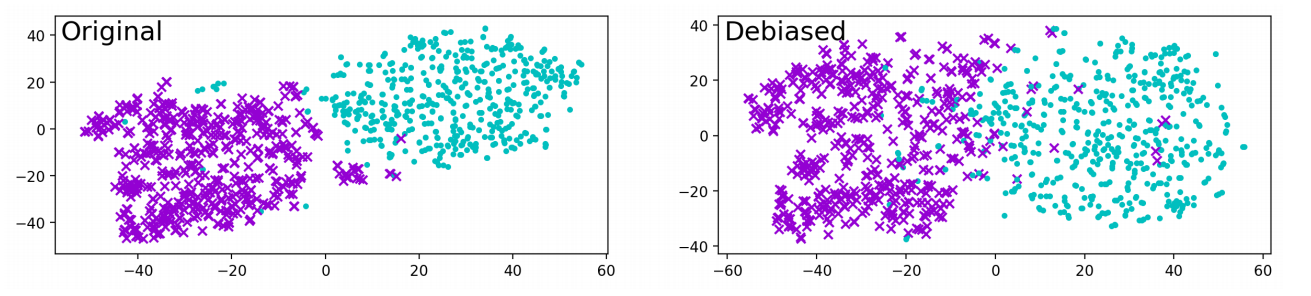}
    \caption{The PCA comparison from \cite{gonen}, showing their results before and after the \emph{Vector Space Transformation} is Applied.}
    \label{PCAGON}
\end{figure}

\begin{figure}
    \centering
    \includegraphics[width=0.4\textwidth]{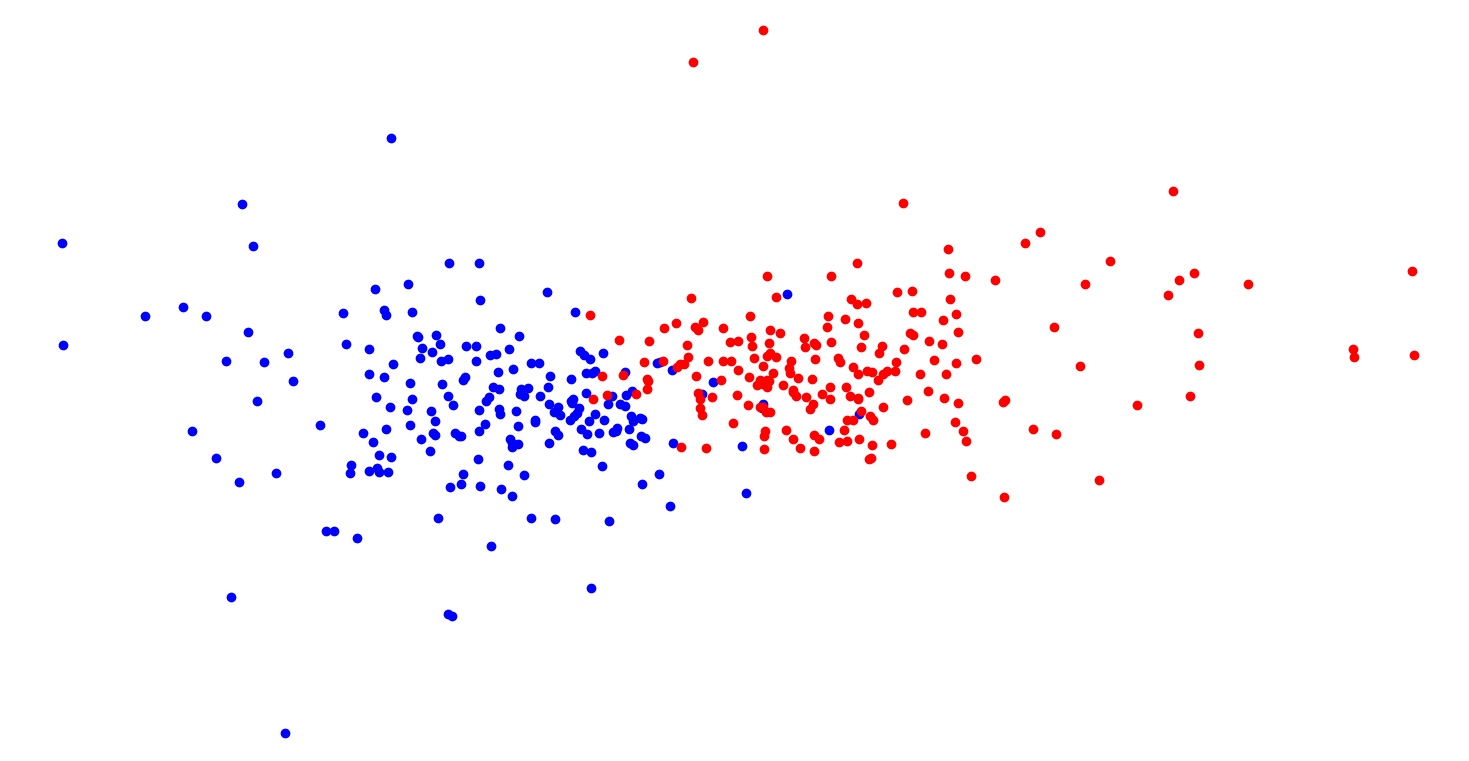}
    \caption{PCA results of the 200 most biased words in each direction from the smaller dataset, prior to any debiasing. The colours distinguish which direction that word is biased in.}
    \label{PCASB}
\end{figure}

\begin{figure}
    \centering
    \includegraphics[width=0.4\textwidth]{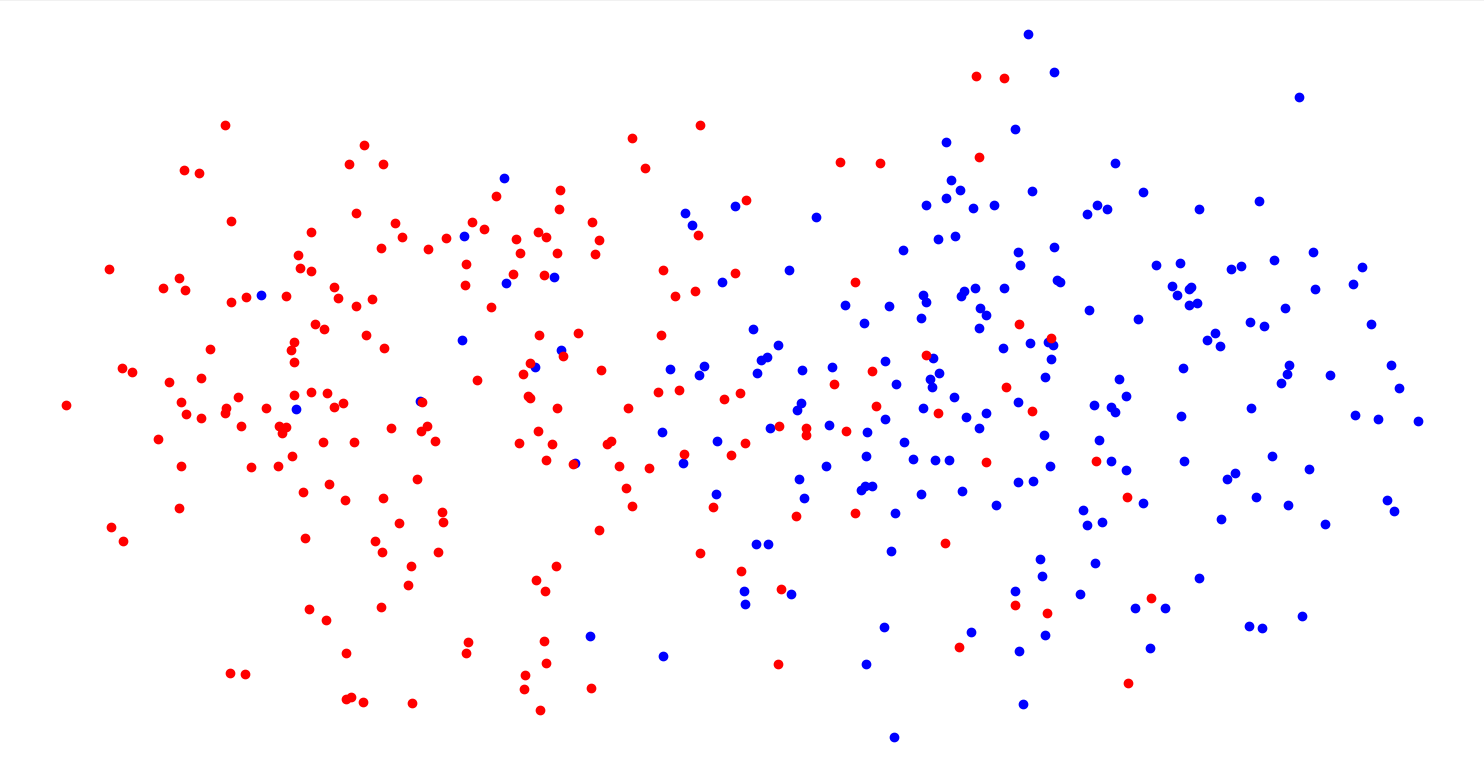}
    \caption{PCA results of the 200 most biased words in each direction from the smaller dataset, after the \emph{Vector Space Transformation} has been applied. The colours distinguish which direction that word is biased in.}
    \label{PCASP}
\end{figure}

\begin{figure}
    \centering
    \includegraphics[width=0.4\textwidth]{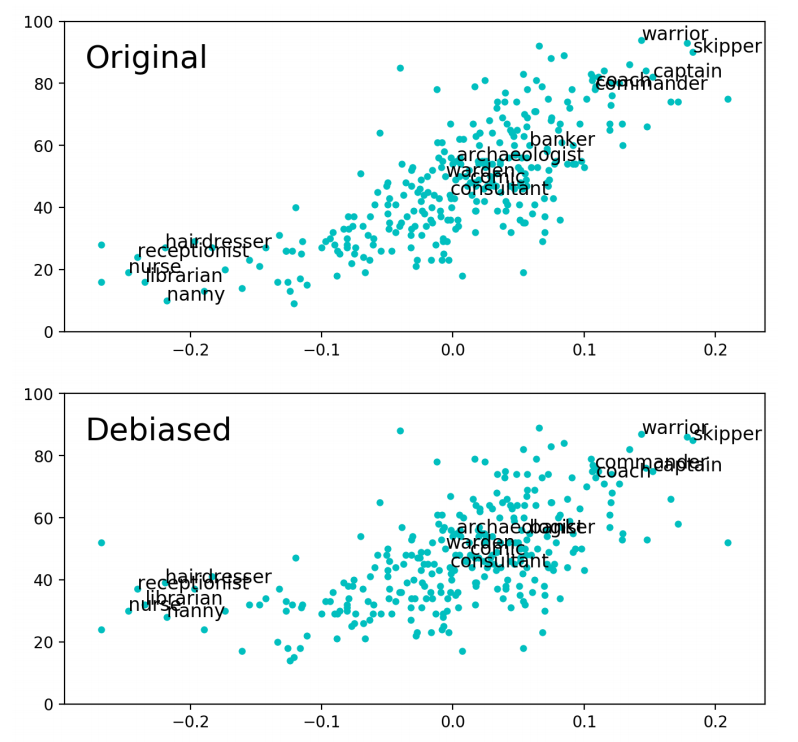}
    \caption{Bias-By-Neighbours plotted against the original bias projection from \cite{gonen}, shown before and after the \emph{Vector Space Transformation} is applied}
    \label{BBNGON}
\end{figure}

\begin{figure}
    \centering
    \includegraphics[width=0.5\textwidth]{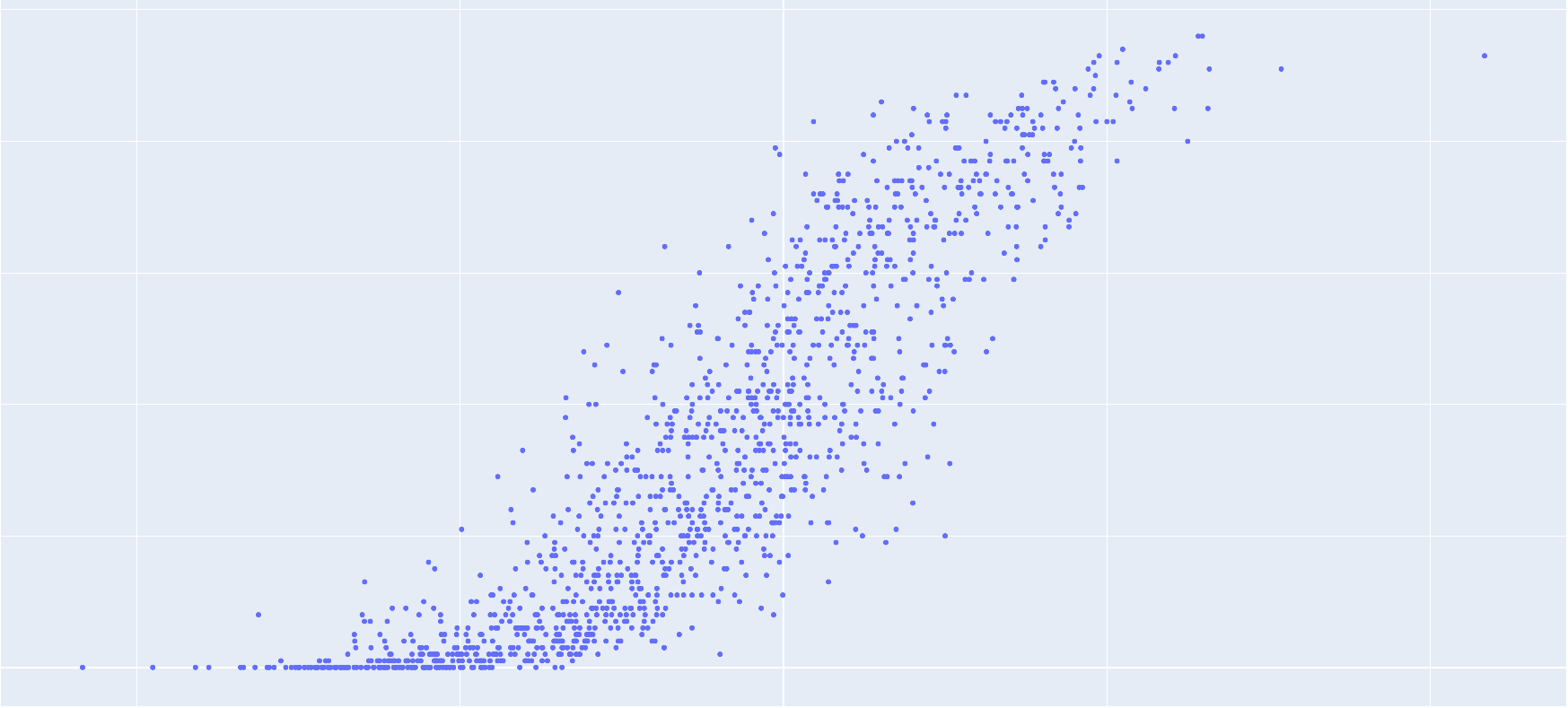}
    \caption{Bias-By-Neighbours plotted against the original bias projection from the smaller dataset, prior to any debiasing.}
    \label{BBNB}
\end{figure}

\begin{figure}
    \centering
    \includegraphics[width=0.5\textwidth]{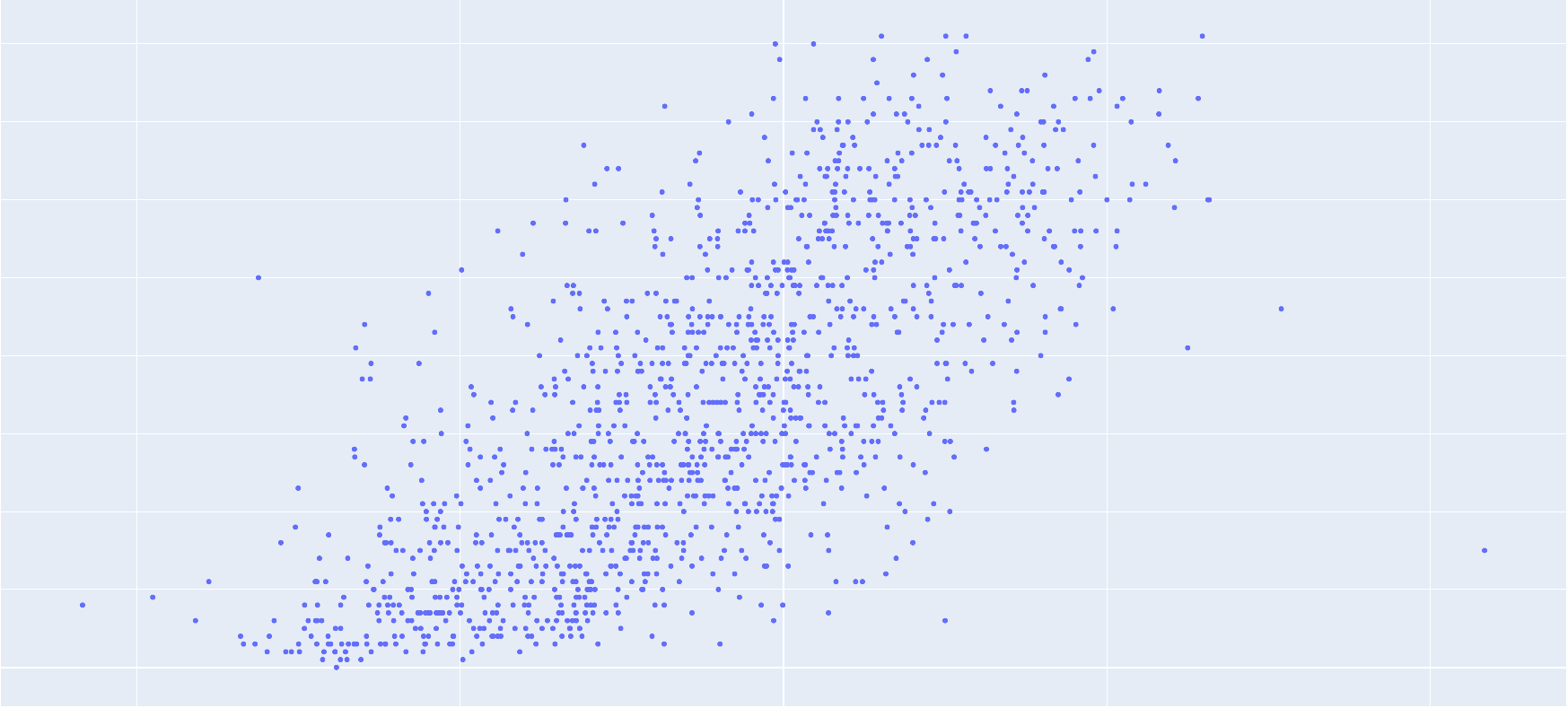}
    \caption{Bias-By-Neighbours plotted against the original bias projection from the smaller dataset, after the \emph{Vector Space Transformation} is applied.}
    \label{BBNP}
\end{figure}

The results indicate that the \emph{Vector Space Transformation} yields a similar reduction in bias for both small and larger results. Table \ref{tab:gonencomp} shows that both K-Means Cluster accuracy and Bias-By-Neighbours Pearson correlation decrease by a small amount after the \emph{Vector Space Transformation} is applied. Plotting the 2 most significant features after PCA, we see that on both datasets the clusters spread out but still remain largely seperate from each other (Figures \ref{PCAGON},\ref{PCASB},\ref{PCASP}). Figures \ref{BBNGON},\ref{BBNB},\ref{BBNA} show that the smaller dataset results in a larger spread but the correlation is still clearly visible.

\clearpage

\subsection{Comparison of Adversarially Debiased results and Linear Transformed results on a smaller dataset}

\begin{table}[]
\resizebox{\textwidth}{!}{\begin{tabular}{lll}
Embedding                                 & K-Means Accuracy & Bias-By-Neighbours Pearson Correlation \\
Non-Debiased                              & 99.4\%           & 0.884                                  \\
Debiased with \emph{Vector Space Transformation} & 88.9\%           & 0.693                                  \\
Debiased with \emph{Adversarial Debiasing}       & 66\%             & 0.431                                  
\end{tabular}}
\caption{A table comparing the results on the K-means clustering test and the Bias-by-Neighbours correlation for the 3 different embeddings I produced.}
\label{tab:12advresults}
\end{table}

\begin{figure}
    \centering
    \includegraphics[width=0.4\textwidth]{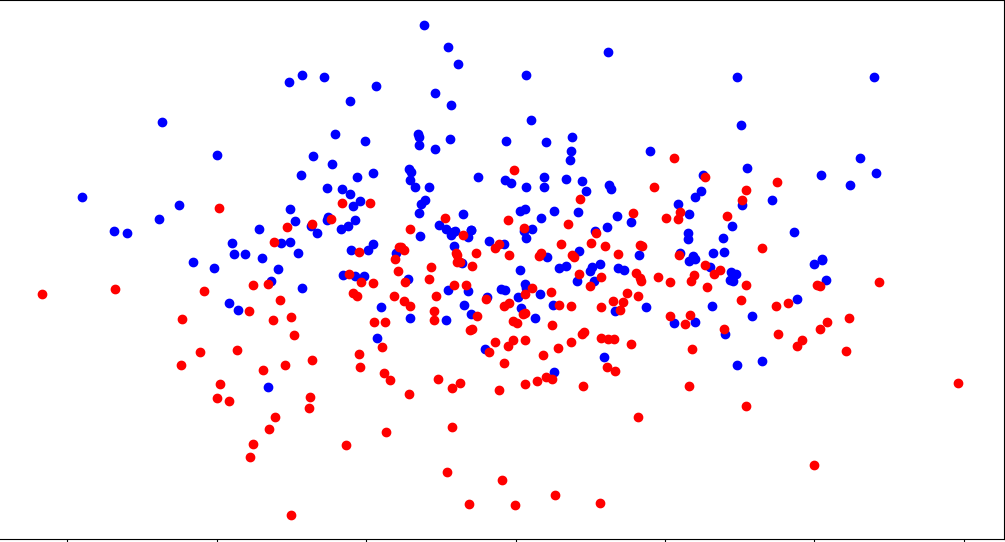}
    \caption{The PCA results of the 200 most biased words after the embedding is adversarially debiased. The colours distinguish which direction that word is biased in.}
    \label{PCAA}
\end{figure}

\begin{figure}
    \centering
    \includegraphics[width=0.5\textwidth]{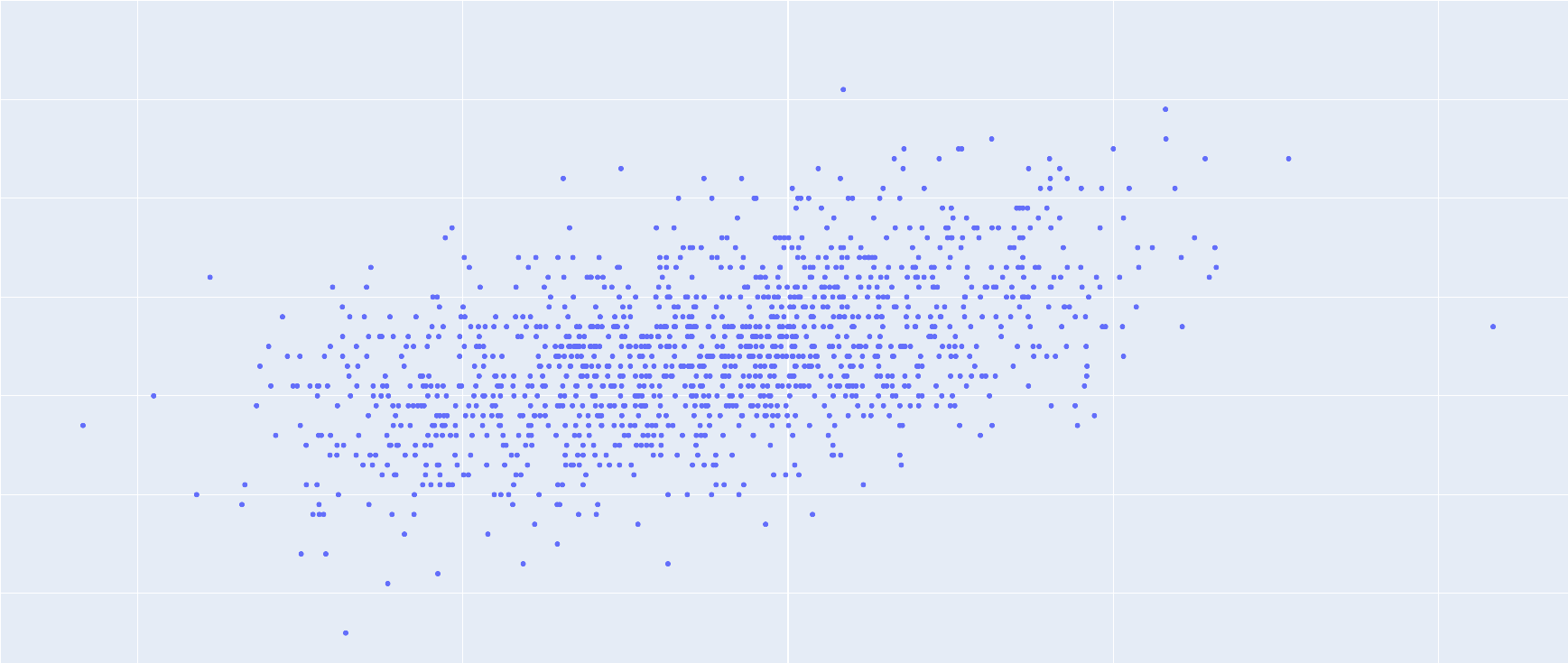}
    \caption{Bias-By-Neighbours plotted against the original bias projection from the smaller dataset, after \emph{Adversarial Debiasing} is applied.}
    \label{BBNA}
\end{figure}

Table \ref{tab:12advresults} compares the results from an embedding that has not been debiased, an embedding debiased with the \emph{Vector Space Transformation} and an embedding debiased with \emph{Adversarial Debiasing}. A K-Means accuracy score of 50\% would mean perfect seperation of word clusters. We see significantly improved results, showing that the most biased words are harder to accurately cluster and the original bias projection is harder to predict from the Bias-By-Neighbours.

The K-Means clustering results are supported by the Adversarially Debiased PCA (Figure \ref{PCAA}) which shows significantly more enmeshed clusters when compared to the \emph{Vector Space Transformation} (Figure \ref{PCASP}). Figure \ref{BBNA} shows significantly less correlation between the current neighbours and original bias projection when compared to a \emph{Vector Space Transformation} (Figure \ref{BBNP}), indicating that the \emph{Adversarial Debiasing} has more effectively removed bias.

When compared using the three experiments shown by \cite{gonen}, it can be seen that \emph{Adversarial Debiasing} has been more effective in comprehensively removing bias than the \emph{Vector Space Transformation}, and the results are not similar.

\clearpage
\section{Conclusion}

I have shown that when experimenting on a smaller dataset, the three expiriments shown in \cite{gonen} (PCA, K-Means Clustering and Bias-By-Nieghbours correlation) all indicate that the \emph{Adversarial Debiasing} is more effective at removing non-superficial bias than the Linear Transformation shown in \cite{bobulaski} . Whilst it is essential to note this was on a significantly smaller dataset, this demonstrates a need for further work into the utility of \emph{Adversarial Debiasing}. A limitation of \emph{Adversarial Debiasing} is the fact it requires the original corpus, not publicly available for the Google News Vectors, and presents a significant training overhead when compared with the \emph{Vector Space Transformation}. \emph{Adversarial Debiasing} may be a useful tool for removing bias, and mitigating harms when these systems are applied in the real world.

Further work should explore if \emph{Adversarial Debiasing} is able to effectively remove bias at scale, and what the applications and limitaitons of the technique are. This work should be informed by the recommendations made by \cite{nlplit1}, which include grounding an understanding of bias in literature outside of NLP and making explicit statements about what harm is caused, and to whom.

\section{Acknowledgments}

This work contains figures from \cite{gonen}, published under the Creative Commons BY-SA 4.0 license \url{https://creativecommons.org/licenses/by-sa/4.0/} which allows for the work to be freely adapted. No figures have been changed.

This work was carried out using the computational facilities of the Advanced Computing Research Centre, University of Bristol - http://www.bristol.ac.uk/acrc/.

Many thanks to Dr. Edwin Simpson, who shared their knowledge and expertise on Natural Language Processing and assisted with the development of this work.
\bibliographystyle{unsrtnat}
\bibliography{references}  






\end{document}